\documentclass[11pt,a4paper]{article}
\usepackage[hyperref]{acl2019}
\usepackage{times}
\usepackage{latexsym}

\usepackage{url}

\usepackage[utf8]{inputenc} 
\usepackage[T1]{fontenc}    
\usepackage{hyperref}       
\usepackage{url}            
\usepackage{booktabs}       
\usepackage{amsfonts}       
\usepackage{microtype}      
\usepackage{float}
\usepackage{amsmath}

\usepackage{xcolor}
\usepackage{subfig}
\usepackage{graphicx}
\usepackage{placeins}

\usepackage{wrapfig}

\newcommand{\cut}[1]{}

\definecolor{boxpink}{RGB}{255,0,255}

\aclfinalcopy
\def\aclpaperid{2118}

\title{Learning to Relate from Captions and Bounding Boxes}

\author{Sarthak Garg\Thanks{ Equal Contribution} , Joel Ruben Antony Moniz\footnotemark[1] , Anshu Aviral\footnotemark[1] , Priyatham Bollimpalli\footnotemark[1] \\
  School of Computer Science \\
  Carnegie Mellon University  \\
  {\tt \{sarthakg, jrmoniz, aanshu, pbollimp\}@cs.cmu.edu}
  }

\begin{document}

\maketitle

\begin{abstract}
    In this work, we propose a novel approach that predicts the relationships between various entities in an image in a weakly supervised manner by relying on image captions and object bounding box annotations as the sole source of supervision. Our proposed approach uses a top-down attention mechanism to align entities in captions to objects in the image, and then leverage the syntactic structure of the captions to align the relations. We use these alignments to train a relation classification network, thereby obtaining both grounded captions and dense relationships. We demonstrate the effectiveness of our model on the Visual Genome dataset by achieving a recall@50 of 15\% and recall@100 of 25\% on the relationships present in the image. We also show that the model successfully predicts relations that are not present in the corresponding captions.
\end{abstract}

\section{Introduction} \label{intro}
Scene graphs serve as a convenient representation to capture the entities in an image and the relationships between them, and are useful in a variety of settings (for example, \citet{johnson2015image, anderson2016spice, liu2017improved}). While the last few years have seen considerable progress in classifying the contents of an image and segmenting the entities of interest without much supervision \citep{he2017mask}, the task of identifying and understanding the way in which entities in an image interact with each other without much supervision remains little explored.

Recognizing relationships between entities is non-trivial because the space of possible relationships is immense, and because there are $\mathcal{O}(n^2)$ relationships possible when $n$ objects are present in an image. On the other hand, while image captions are easier to obtain, they are often not completely descriptive of an image \citep{krishnavisualgenome}. Thus, simply parsing a caption to extract relationships from them is likely to not sufficiently capture the rich content and detailed spatial relationships present in an image. 

Since different images have different objects and captions, we believe it is possible to get the information that is not present in the caption of one image from other similar images which have the same objects and their captions. In this work, we thus aim to learn the relationships between entities in an image by utilizing only image captions and object locations as the source of supervision. Given that generating a good caption in an image requires one to understand the various entities and the relationships between them, we hypothesize that an image caption can serve as an effective weak supervisory signal for relationship prediction.

\begin{figure*}
\centering
\includegraphics[width=0.8\textwidth]{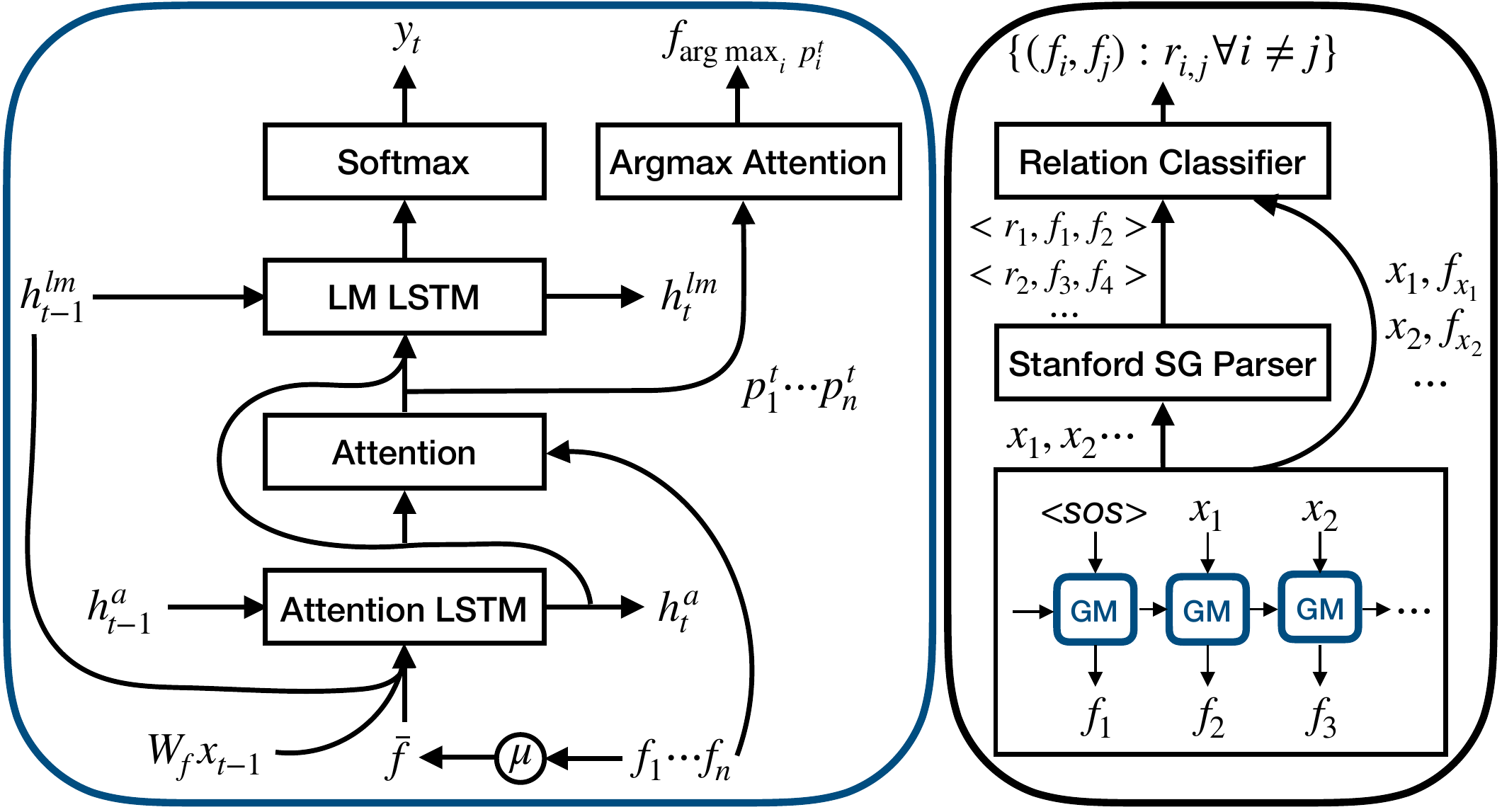}
\caption{C-GEARD Architecture (left) and it's integration with the relation classifier (right). C-GEARD acts as the Grounding Module (GM) in our relation classifier.}
\label{fig:cgeard_model}
\end{figure*}

\section{Related work}  \label{related}
The task of Visual Relationship Detection has been the main focus of several recent works \citep{lu2016visual, li2017vip, zhang2017visual, dai2017detecting, hu2017modeling, liang2017deep, yin2018zoom}. The goal is to detect a generic $<$subject, predicate, object$>$ triplet present in an image. Various techniques have been proposed to solve this task, such as \cut{by modeling relations as the interactions between different objects in an image are initially modeled in the form of visual phrases \citep{desai2012detecting, choi2013understanding}, verbs \cite{fader2011identifying, chao2015hico, gupta2008beyond} and actions \cite{gupta2009observing, yao2010modeling};} by using language priors \cite{lu2016visual, yatskar2016situation}, deep network models \cite{zhang2017visual, dai2017detecting, zhu2018deep, yin2018zoom}, referring expressions \cite{hu2017modeling, cirik2018using} and reinforcement learning \cite{liang2017deep}. Recent work has also studied the closely related problem of Scene Graph Generation, \citep{li2017scene, newell2017pixels, xu2017scene, yang2017support, yang2018graph}. The major limitation of the aforementioned techniques is that they are supervised, and require the presence of ground truth scene graphs or relation annotations. Obtaining these annotations can be an extremely tedious and time consuming process that often needs to be done manually. Our model in contrast does the same task through weak supervision, which makes this annotation significantly easier.

Most similar to our current task is work in the domain of Weakly Supervised Relationship Detection. \citet{DBLP:journals/corr/PeyreLSS17} uses weak supervision to learn the visual relations between the pairs of objects in an image using a weakly supervised discriminative clustering objective function \citep{bach2008diffrac}, while \citet{zhang2017ppr} uses a region-based fully convolutional network neural network to perform the same task. They both use $<$subject, predicate, object$>$ annotations without any explicit grounding in the images as the source of weak supervision, but require these annotations in the form of image-level triplets. Our task, however, is more challenging, because free-form captions can potentially be both extremely unstructured and significantly less informative than annotated structured relations.

\section{Proposed Approach} \label{problem}
Our proposed approach consists of three sequential modules: a feature extraction module, a grounding module and a relation classifier module. Given the alignments found by the grounding module, we train the relation classifier module, which takes in a pair of object features and classifies the relation between them.
\subsection{Feature Extraction}
Given an image $I$ with $n$ objects and their ground truth bounding boxes \{$b_1, b_2, \ldots, b_n$\}, the feature extraction module extracts their feature representations $\mathcal{F} = \{f_1, f_2, \ldots, f_n\}$. To avoid using ground truth instance-level class annotations that would be required to train an object detector, we use a ResNet-152 network pre-trained on ImageNet as our feature extractor. For every object $i$, we crop and resize the portion of the image $I$ corresponding to the bounding box $b_i$ and feed it to the ResNet model to get its feature representation $f_i$. $f_i$ is a dense $d$-dimensional vector capturing the semantic information of the $i^{th}$ object. Note that we do not fine-tune the ResNet architecture.

\subsection{Grounding Caption Words to Object Features}
\label{sec:grounding}
Given an image $I$, its caption consisting of words $\mathcal{W} = $ \{$w_1, w_2 \ldots, w_k$\} and the feature representations $\mathcal{F}$ obtained above, the grounding module aligns the entities and relations found in the captions with the objects' features and the features corresponding to pairs of objects in the image. It thus aims to find the subset of words in the caption corresponding to entities $\mathcal{E} \subseteq \mathcal{W} \mid \mathcal{E} = \{e_{i_1}, e_{i_2}, \ldots, e_{i_m}\}$, and to ground each such word with its best matching object feature $f_{i_j}$.
It also aims to find the subset of relational words $\mathcal{R} \subseteq \mathcal{W} \mid \mathcal{R} = \{r_{i_1}, r_{i_2}, \dots, r_{i_l}\}$ and to ground each relation to a pair of object features \{$f_{i, subj}, f_{i, obj}$\} which correspond to the subject and object of that relation.

To identify and ground the relations between entities in an image, we propose C-GEARD (Captioning-Grounding via Entity Attention for Relation Detection). C-GEARD passes the caption through the Stanford Scene Graph Parser \citep{schuster2015generating} to get the set of triplets $\mathcal{T} = \{(s_1, p_1, o_1), (s_2, p_2, o_2), \ldots, (s_g, p_g, o_g)\}$. Each triplet corresponds to one relation present in the caption. For $(s_i, p_i, o_i) \in \mathcal{T}$, $s_i$, $p_i$ and $o_i$ denote  \textit{subject}, \textit{predicate} and \textit{object} respectively. The entity and relation subsets are then constructed as:
\begin{gather*}
\mathcal{E} = \bigcup_{(s_i, p_i, o_i) \in \mathcal{T}}\{s_i, o_i\} \;\;\;\;\;\;
\mathcal{R} = \bigcup_{(s_i, p_i, o_i) \in \mathcal{T}} \{p_i\}
\end{gather*}
Captioning using visual attention has proven to be very successful in aligning the words in a caption to their corresponding visual features, such as in \citet{anderson2018bottom}. As shown in Figure \ref{fig:cgeard_model}, we adopt the two-layer LSTM architecture in \citet{anderson2018bottom}; our end goal, however, is to associate each word with the closest object feature rather than producing a caption. 

The lower Attention LSTM cell takes in the words and the global image context vector ($\bar{f}$, the mean of all features $\mathcal{F}$), and its hidden state $h_{t}^{a}$ acts as a query vector. This query vector is used to attend over the object features $\mathcal{F}=$ \{$f_1, f_2, \ldots, f_n$\} (serving as both key and value vectors) to produce an attention vector which summarizes the key visual information needed for predicting the next word. The Attention module is parameterized as in \citet{bahdanau2014neural}. The concatenation of the query vector and the attention vector is passed as an input to the upper LM-LSTM cell, which predicts the next word of the caption.

The model is trained by minimizing the standard negative log-likelihood loss.
\begin{equation*}
\mathcal{L}_{NLL} = -\frac{1}{k}\sum_{i=1}^{k}\log(\mathcal{P}(w_i | w_1 \ldots w_{i - 1}))
\end{equation*}

Let $p^{w_i}_{x}$ denote the attention probability over feature $x$ when \textit{previous word $w_{i - 1}$} is fed into the LSTM. C-GEARD constructs alignments of the entity and relation words as follows:

\begin{align*}
 w_e & \xrightarrow[]{\text{align}} \underset{f \in \mathcal{F}}{\arg\max} (p^{w_e}_{f}) \; \; \; \forall w_e \in \mathcal{E}\\
  p_i & \xrightarrow[]{\text{align}} (\underset{f \in \mathcal{F}}{\arg\max} (p^{s_i}_{f}), \underset{f \in \mathcal{F}}{\arg\max} (p^{o_i}_{f})) \; \; \; \forall p_i \in \mathcal{R}
\end{align*}

\subsection{Relation Classifier} 
We run the grounding module C-GEARD over the training captions to generate a ``grounded'' relationship dataset consisting of tuples \{$((f_i, f_j), p_{i, j})$\}, where $f_i$ and $f_j$ are two object features and $p_{i, j}$ refers to the corresponding aligned predicates. These predicates occur in free form; however, the relations in the test set are restricted to only the top $50$ relation classes. We manually annotate the correspondence between the $300$ most frequent parsed predicates and their closest relation class. For example, we map the parsed predicates \textit{dress in}, \textit{sitting in} and \textit{inside} to the canonical relation class \textit{in}. Using this mapping we get tuples of the form \{$((f_i, f_j), c_{i, j})$\} where $c_{i, j}$ denotes the canonical class corresponding to $p_{i, j}$ 

Since this dataset is generated by applying the grounding module on the set of all images and the corresponding captions, it pools the relation information from across the whole dataset, which we then use to train our relation classifier.

We parameterize the relation classifier with a $2$-layer MLP. Given the feature vectors of any two objects $f_i$ and $f_j$, the relation classifier is trained to classify the relation $c_{i, j}$ between them.
\subsection{Model at Inference}
During inference, the features extracted from each pair of objects is passed through the relation classifier to predict the relation between them.

\begin{table*}[t]
\setlength\tabcolsep{3.2pt}
\centering
\begin{tabular}{|c|c|c|c||c|c|}
\hline
Recall@ & \begin{tabular}[c]{@{}c@{}}IMP \\ \cite{xu2017scene}\end{tabular} & \begin{tabular}[c]{@{}c@{}}Pixel2Graph \\ \cite{newell2017pixels} \end{tabular} & \begin{tabular}[c]{@{}c@{}}Graph-RCNN \\ \cite{yang2018graph}\end{tabular} & \begin{tabular}[c]{@{}c@{}}Parsed caption \\ (baseline)\end{tabular} & \begin{tabular}[c]{@{}c@{}}C-GEARD \\ (ours)\end{tabular} \\ \hline
50 & 44.8 & 68.0 & 54.2 & 4.1 & 15.3 \\ \hline
100 & 53.0 & 75.2 & 59.1 & 4.1 & 25.2 \\ \hline
\end{tabular}
\caption{Comparison with respect to Recall@50 and Recall@100 on PredCls metric, in \%.}
\label{table:results}
\end{table*}

\section{Experiments}
\label{experiments}
\subsection{Dataset} 
We use the MS COCO \cite{DBLP:journals/corr/LinMBHPRDZ14} dataset for training and the Visual Genome \cite{krishnavisualgenome} dataset for evaluation. MS COCO has images and their captions, and Visual Genome contains images and their associated scene graphs. The Visual Genome dataset consists in part of MS COCO images, and since we require ground truth captions and bounding boxes during training, we filter the Visual Genome dataset by considering only those images which are part of the original MS COCO dataset. Similar to \citet{xu2017scene}, we manually remove poor quality and overlapping bounding boxes with ambiguous object names, and filter to keep the 150 most frequent object categories and 50 most frequent predicates. Our final dataset thus comprises of 41,731 images with 150 unique objects and 50 unique relations. We use a 70-30 train-validation split. We use the same test set as \citet{xu2017scene}, so that the results are comparable with other supervised baselines. 

\subsection{Baselines} 
Since, to the best of our knowledge, this work is the first to introduce the task of weakly supervised relationship prediction solely using captions and bounding boxes, we do not have any directly comparable baselines, i.e., all other work is either completely supervised or relies on all ground truth entity-relation triplets being present at train time. Consequently, we construct baselines relying solely on captions and ground truth bounding box locations that are comparable to our task. In particular, running the Stanford Scene Graph Parser \cite{schuster2015generating} on ground truth captions  constructs a scene graph just from the image captions (which almost never capture all the information present in an image). We use this baseline as a lower bound, and to obtain insight into the limitations of scene graphs directly generated from captions. On the other hand, we use supervised scene graph generation baselines \cite{yang2018graph, newell2017pixels} to upper bound our performance, since we rely on far less information and data. 

\subsection{Evaluation Metric} 
As our primary objective is to detect relations between entities, we use the PredCls evaluation metric \citep{xu2017scene}, defined as the performance of recognizing the relation between two objects given their ground truth locations. We only use the entity bounding boxes' locations without knowing the ground truth objects they contain. We show results on Recall@$k$ (the fraction of top $k$ relations predicted by the model contained in the ground truth) for $k=50$ and $100$. The predicted relations are ranked over all objects pairs for all relation classes by the relation classifier's model confidence.

\begin{figure*}[!htb]
\centering
\minipage[t]{0.8\textwidth}
  \includegraphics[width=0.98\linewidth]{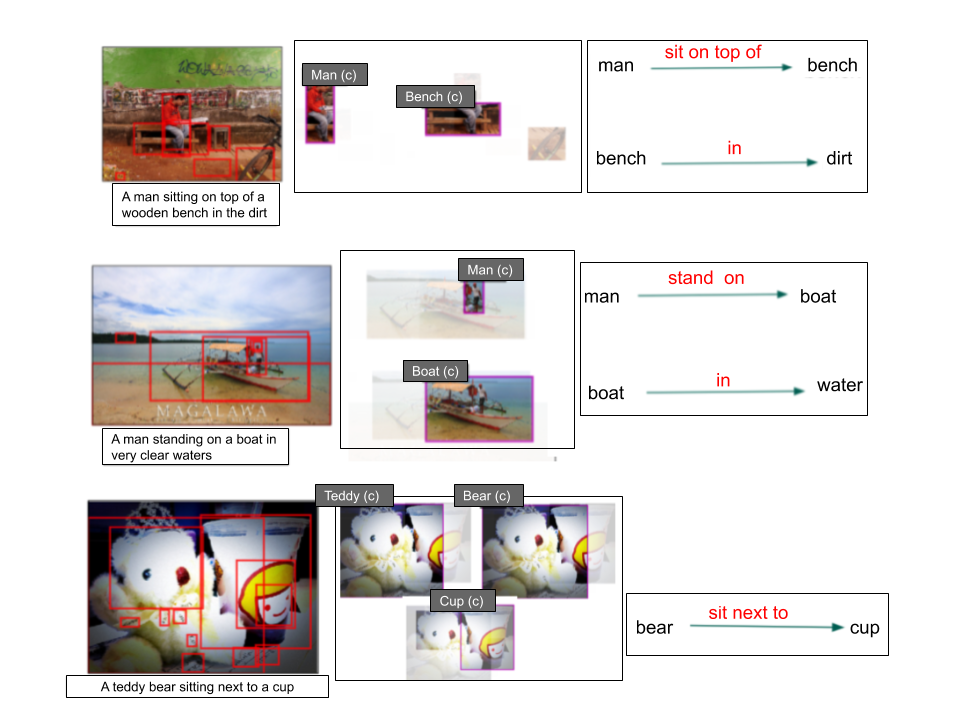}
  \caption{Attention masks for each of the entities in the caption for C-GEARD. The output of the Stanford Scene Graph Parser is given on the right.}\label{fig:cgeard_vis}
\endminipage\hfill
\end{figure*}

\section{Results and Discussion}
\label{results}
\subsection{Performance}
We show the performance of C-GEARD in Table~\ref{table:results}. We compare its performance with various supervised baselines, as well as a baseline which parses relations from just the caption using Stanford Scene Graph Parser \cite{schuster2015generating} (caption-only baseline), on the PredCls metric. Our proposed method substantially outperforms the caption-only baseline. This shows that our model predicts relationships more successfully than by purely relying on captions, which contain limited information. This in turn supports our hypothesis that it is possible to detect relations by pooling information from captions across images, without requiring all ground truth relationship annotations for every image.

Note that our model is at a significant disadvantage when compared to supervised approaches. First, we use pre-trained ResNet features (trained on a classification task) without any fine-tuning; supervised methods, however, use Faster RCNN \cite{ren2015faster}, whose features are likely much better suited for multiple objects. Second, supervised methods likely have a better global view than C-GEARD, because Faster RCNN provides a significantly larger number of proposals, while we rely on ground truth regions which are far fewer in number. Third, and most significant, we have no ground truth relationship or class information, relying purely on weak supervision from captions to provide this information. Finally, since we require captions, we use significantly less data, training on the subset of Visual Genome overlapping  with MS COCO (and has ground truth captions as a result).

\subsection{Relation Classification}
We train the relation classifier on image features of entity pairs and using the relations found in the caption as the only source of supervision. On the validation set, we obtain a relation classification accuracy of 22\%.

We compute the top relations that the model gets most confused about, shown in Table \ref{table:confusion}. We observe that even when the predictions are not correct, they are semantically close to the ground truth relation class.

\begin{table}[!h]
\centering
\begin{tabular}{|c|c|}
\hline
Relation & Confusion with Relations \\
\hline
above & on, with, sitting on, standing on, of \\
\hline
carrying & holding, with, has, carrying, on \\
\hline
laying on & on, lying on, in, has \\
\hline
mounted on & on, with, along, at, attached to \\
\hline
\end{tabular}
\caption{Relations the classification model gets most confused about}
\label{table:confusion}
\end{table}
\subsection{Visualizations}
Three images with their captions are given in Figure~\ref{fig:cgeard_vis}. We can see that C-GEARD generates precise entity groundings, and that the Stanford Scene Graph Parser generates correct relations. This results in the correct grounding of the entities and relations which yields accurate training samples for the relation classifier.

\section{Conclusion}
\label{conclusion}

In this work, we propose a novel task of weakly-supervised relation prediction, with the objective of detecting relations between entities in an image purely from captions and object-level bounding box annotations without class information. 
Our proposed method builds upon top-down attention \citep{anderson2018bottom}, which generates captions and grounds word in these captions to entities in images. We leverage this along with structure found from the captions by the Stanford Scene Graph Parser \citep{schuster2015generating} to allow for the classification of relations between pairs of objects without having ground truth information for the task. Our proposed approaches thus allow weakly-supervised relation detection.

There are several interesting avenues for future work. One possible line of work involves removing the requirement of ground truth bounding boxes altogether by leveraging a recent line of work that does weakly-supervised object detection (such as \citep{oquab2015object, bilen2016weakly, zhang2018zigzag, bai2017multiple, arun2018dissimilarity}). This would reduce the amount of supervision required even further. An orthogonal line of future work might involve using a Visual Question Answering (VQA) task (such as in \citet{krishnavisualgenome}), either on its own replacing the captioning task, or in conjunction with the captioning task with a multi-task learning objective.

\section*{Acknowledgements} 
We would like to thank Louis-Philippe Morency, Carla Viegas, Volkan Cirik and Barun Patra for helpful discussions and feedback. We would also like to thank the anonymous reviewers for their insightful comments and suggestions.

\bibliographystyle{acl_natbib}
\bibliography{main}

\clearpage
\newpage
\appendix
\section{Appendix}
\label{appendix}
\subsection{Failure Cases when Relying Solely on Captions}
\begin{figure}[!htb]
\minipage{0.49\textwidth}
  \includegraphics[width=\linewidth]{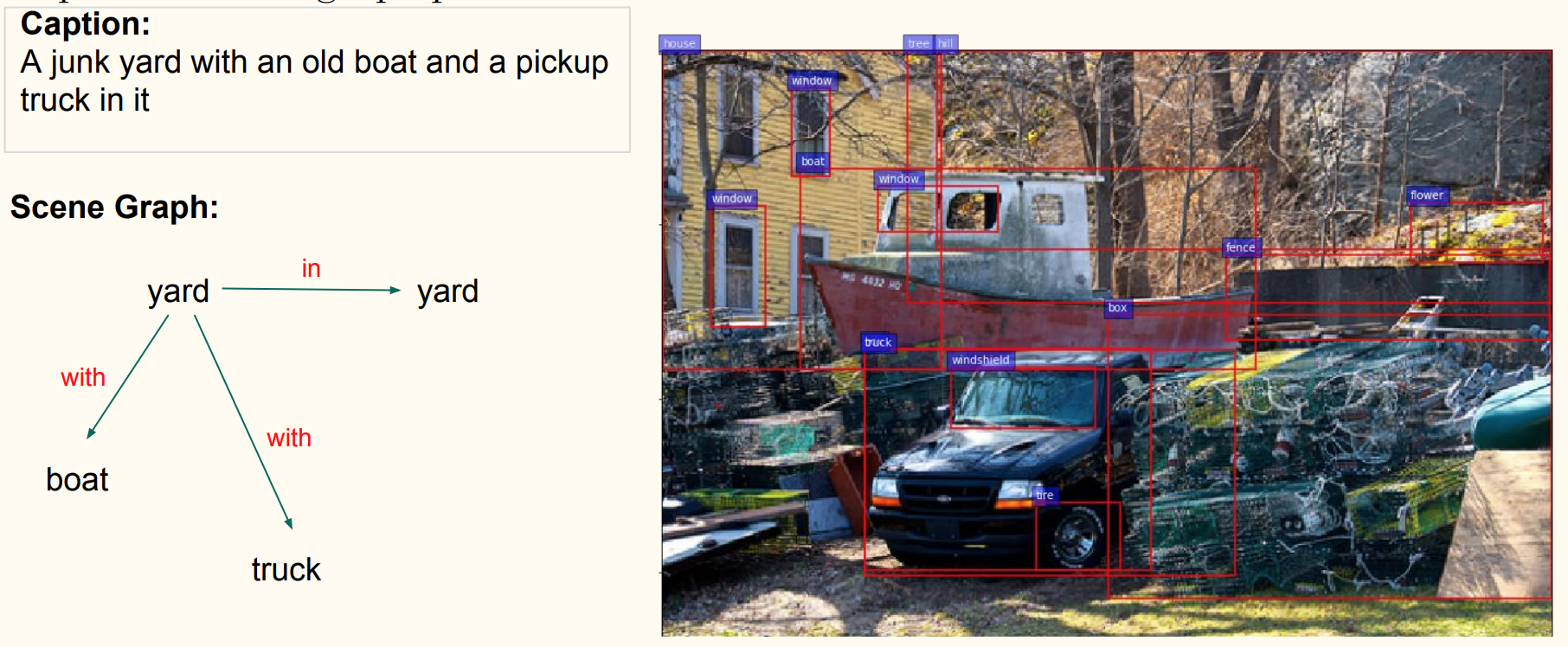}
  \caption{Failure case where the scene graph parser makes errors}\label{fig:fail1}
\endminipage\hfill
\minipage{0.49\textwidth}
  \includegraphics[width=\linewidth]{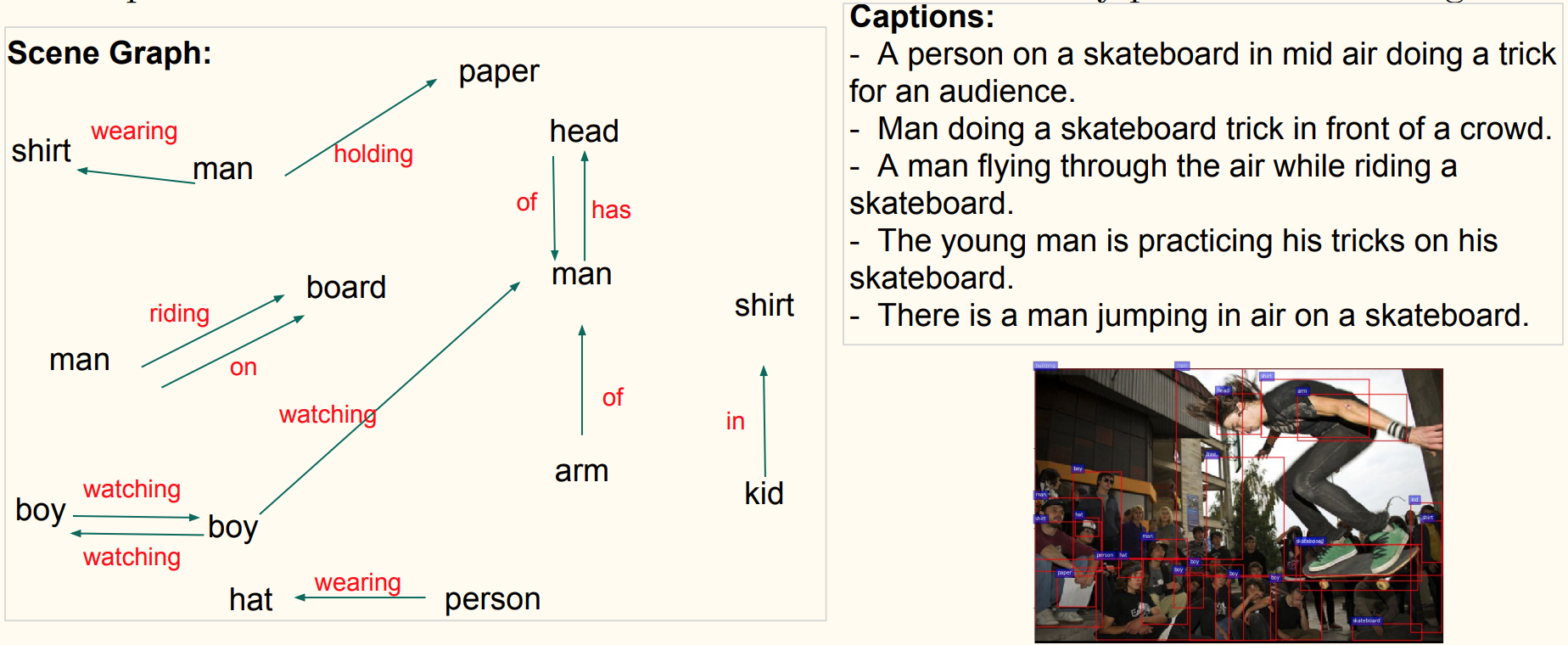}
  \caption{Failure case where all captions are insufficiently descriptive}\label{fig:fail2}
\endminipage\hfill\minipage{0.49\textwidth}
  \includegraphics[width=\linewidth]{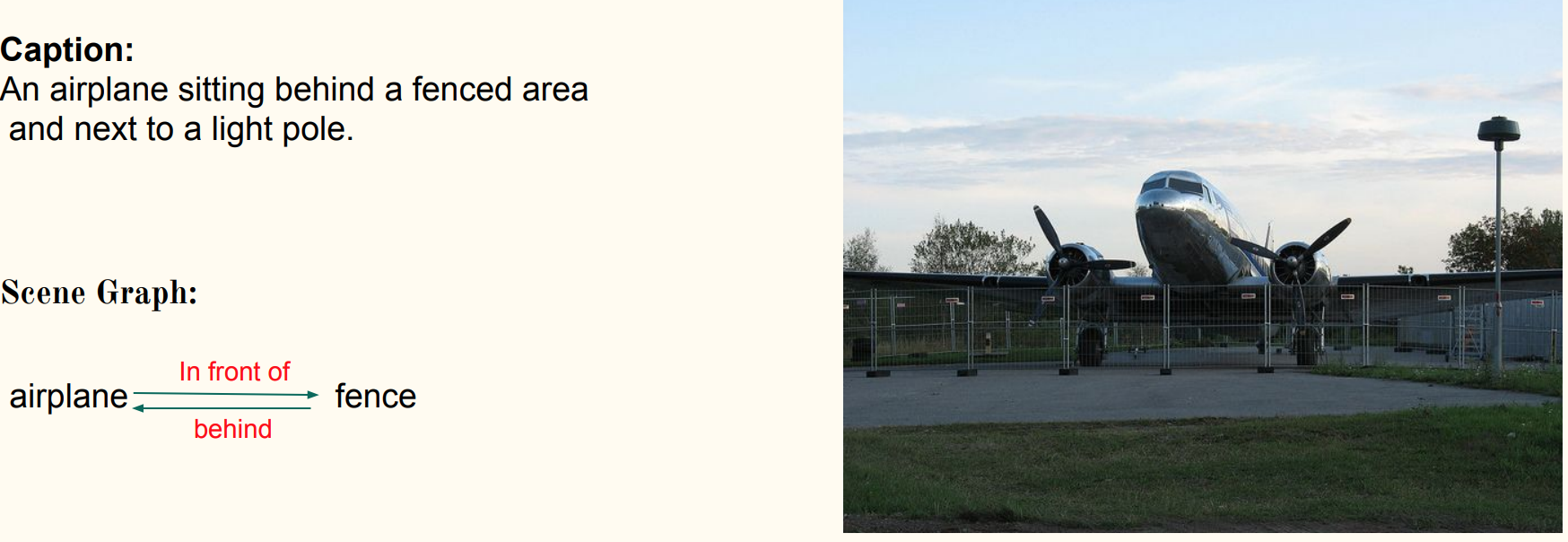}
  \caption{Failure case where the caption does not capture all relationships}\label{fig:fail3}
\endminipage\hfill
\end{figure}

In this section, we identify the key failure cases when relying solely on captions. These failures are primarily due to scene graph parser errors, insufficient information present in captions, and the inability of captions to capture all relationships present in the image.

\subsubsection{Scene Graph Parser Errors}
Generally, the scene graph parser is as effective as using human-constructed scene graphs \citep{schuster2015generating}. However, there exist cases where the scene graph generated from the caption by the Stanford Scene Graph Parser is incorrect. For instance, in Figure~\ref{fig:fail1}, the parser yields two ``yard'' nodes. However, we observe that the majority of the errors are caused by the two subsequently described issues.

\subsubsection{Insufficient Caption Information}
We find that captions describe far less information than actually present in the image. For example, in Figure~\ref{fig:fail2}, though there are multiple objects and relations in the image, none of the five captions are able to completely capture everything in the image.

\subsubsection{Unable to Capture All Relationships}
We find that captions don’t adequately capture all relationships. For example, there are multiple relations such as beneath-behind and up-upward that are not correctly captured. In other cases, some relations actually present in the image are missing-- one such  example the inability to capture transitive relations. For example, in Figure~\ref{fig:fail3}, while the caption indicates that the light pole is next to the airplane, and that the airplane is behind a fence, the caption fails to capture the transitivity (i.e., that the light pole is behind the fence).

\subsection{Correlation between Captions and Ground Truth}

\begin{figure}[!htb]
\minipage{0.49\textwidth}
  \includegraphics[width=\linewidth]{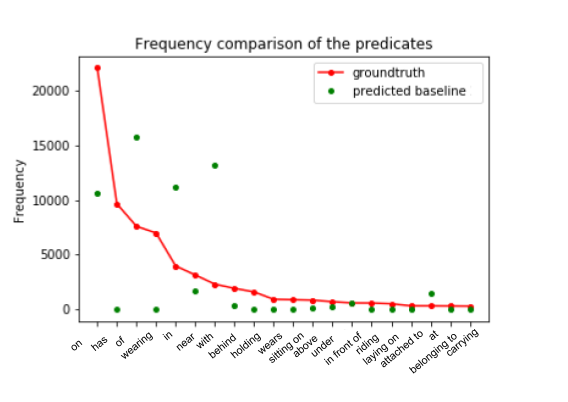}
  \caption{Correlation between the ground truth triplets and the triplets present in captions}\label{fig:rel_freq}
\endminipage\hfill
\end{figure}

Our model formulation aims to pool information about subject-predicate-object triplets from the entire corpus of captions, and to use it to densely identify relations between entities in a single image. To validate whether the most common ground truth relation classes are actually present in the captions, we use the Stanford Scene Graph Parser to extract the predicates and compare their frequency counts with the ground truth relations. This correlation is demonstrated in Fig \ref{fig:rel_freq}.

\begin{figure*}[h]
    \centering
     \includegraphics[width=0.8\textwidth]{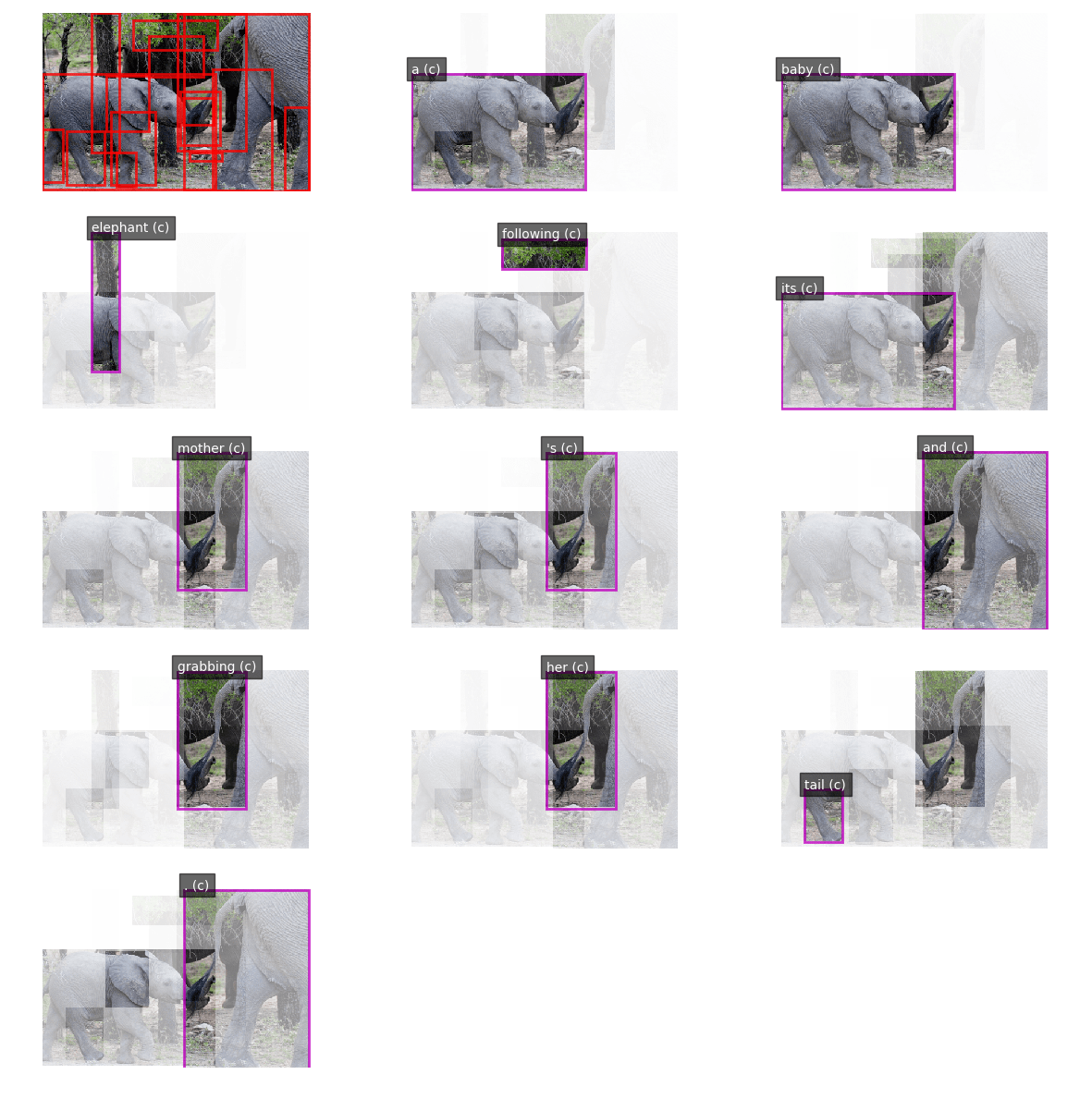}
    \caption{Examples of failures of entity and relation groundings generated by C-GEARD}
    \label{fig:badviz2}
\end{figure*}

\subsection{Failure to Ground Cluttered Scenes}

One failure case for C-GEARD is shown in Figure~\ref{fig:badviz2}. The main reason for this is the large number of ground truth bounding boxes present in the image, which led to the model being unable to correctly capture the groundings.

\subsection{Importance of ResNet Features}

We tried two variants of extracting ResNet features given ground-truth bounding boxes. In the first, we used a fully convolutional approach, using the original object sizes. However, we observed extremely poor performance, and hypothesize that classification networks trained on ImageNet are tuned to ignore small objects. To resolve this, we resized objects so that their larger side is of size 224. We observed significantly better performance; consequently, all reported numbers use these features.

To validate that the benefits observed were due to the changed object feature representations, we trained a simple classifier using the 50 VG object classes with a linear layer (we tried other variants as well, but all other results obtained were comparable). We observed a substantial difference in the performance between these two variants: 45\% accuracy vs 54\% respectively.

\subsection{Hyperparameters}

We train the top-down attention model with entity attention dimension of 512, tanh non-linearity and batch size of 100. We used both the language model LSTM and the attention LSTM with 1000 hidden cells. 
The ResNet extracted object features were 2048 dimensional and the word embeddings were initialized to FastText embeddings of 300 dimensions. Finally, we train our model using an Adam optimizer and a learning rate of 0.0001 for 75 epochs. We train the relation classifier using a simple MLP with 2 hidden layers of 64 units each, with dropout of 0.5 using Adam optimizer and learning rate of 0.001 for 50 epochs.

\subsection{Model Training and Inference}

\begin{figure*}[!bht]
    \centering
     \includegraphics[height=0.96\textheight]{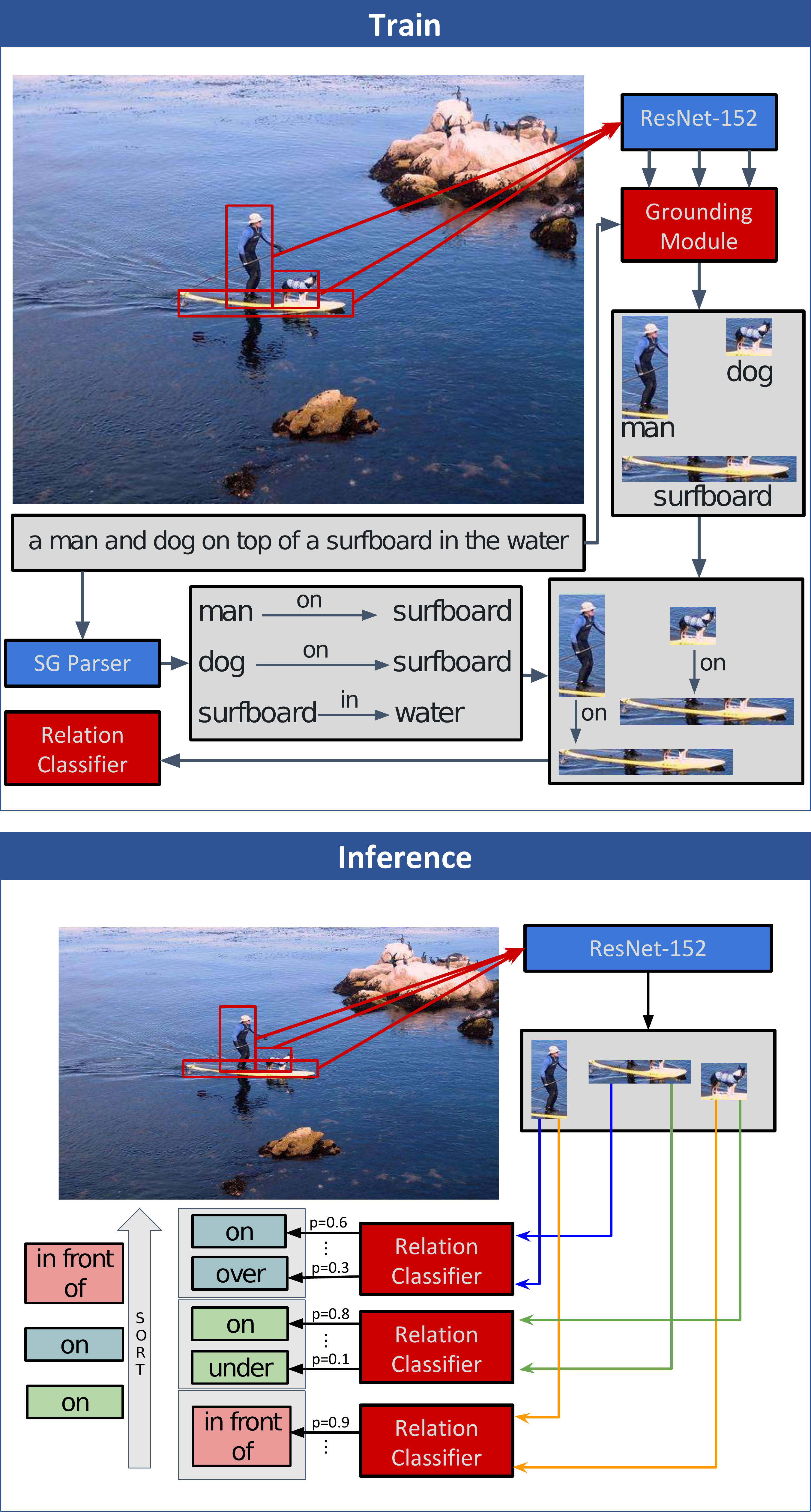}
    \caption{C-GEARD's training and inference.}
    \label{fig:train_infer_flow}
\end{figure*}

Figure~\ref{fig:train_infer_flow} visually explains the training and inference of our proposed C-GEARD architecture.

\end{document}